\documentclass[11pt,a4paper]{article}
\usepackage[hyperref]{emnlp2018}
\usepackage{times}
\usepackage{latexsym}
\usepackage{algorithm2e}
\usepackage{paralist}
\usepackage{graphicx}
\usepackage{hyperref}

\graphicspath{ {./images/} }

\usepackage{url}

\aclfinalcopy

\title{Embedding Individual Table Columns for Resilient SQL Chatbots}

\author{
Bojan Petrovski$^{\dagger1}$,
Ignacio Aguado$^{\ddagger1}$,
Andreea Hossmann$^\ddagger$,
Michael Baeriswyl$^\ddagger$,
Claudiu Musat$^\ddagger$
\\ 
$\dagger$ School of Computer and Communication Sciences, EPFL, Switzerland\\
$\ddagger$ Artificial Intelligence Group - Swisscom AG\\
firstname.lastname@\{epfl.ch, swisscom.com\}
}

\date{}

\begin{document}
\maketitle
\begin{abstract}
Most of the world's data is stored in relational databases. Accessing these requires specialized knowledge of the Structured Query Language (SQL), putting them out of the reach of many people. A recent research thread in Natural Language Processing (NLP) aims to alleviate this problem by automatically translating natural language questions into SQL queries. While the proposed solutions are a great start, they lack robustness and do not easily generalize: the methods require high quality descriptions of the database table columns, and the most widely used training dataset, WikiSQL, is heavily biased towards using those descriptions as part of the questions.

In this work, we propose solutions to both problems: we entirely eliminate the need for column descriptions, by relying solely on their contents, and we augment the WikiSQL dataset by paraphrasing column names to reduce bias. We show that the accuracy of existing methods drops when trained on our augmented, column-agnostic dataset, and that our own method reaches state of the art accuracy, while relying on column contents only.
\end{abstract}

\section{Introduction} 
\label{Introduction}

\footnotetext[1]{equal contribution}

Recent developments in Natural Language Understanding (NLU) have led to a big proliferation of text- and speech-based bot interfaces. Home appliances, such as smart speakers and chatbots, rely mostly on a well-structured knowledge base or an external Application Programming Interface (API) to provide the desired response. This limits the usability of such systems in a context where the data is stored in a (local) relational database.

This constraint led to the development of text to Structured Query Language (SQL) systems, also known as SQL bots. Given a question, in natural language, pertaining to a certain database table, these bots will automatically generate the corresponding SQL query and return the requested data. Considering the vast usage of relational databases on the internet and in private companies, SQL bots are a simple new interface that enables non-technical people to access data. 

The first approaches in the field relied on parsers and pattern-matching rules to understand the question and produce appropriate answers \cite{DBLP}. Later developments introduced semantic grammar systems and intermediate language systems \cite{DBLP}. More recently, new NLU methods, such as pointer-networks, pushed the state-of-the-art results in several domains, including parsing \cite{NIPS2015_5866}. Current state-of-the-art models are based on sketches and have primarily two inputs: the question and the descriptions of the table columns (i.e., the column names). 

\begin{figure*}[]

\centering
\includegraphics[width=0.65\textwidth]{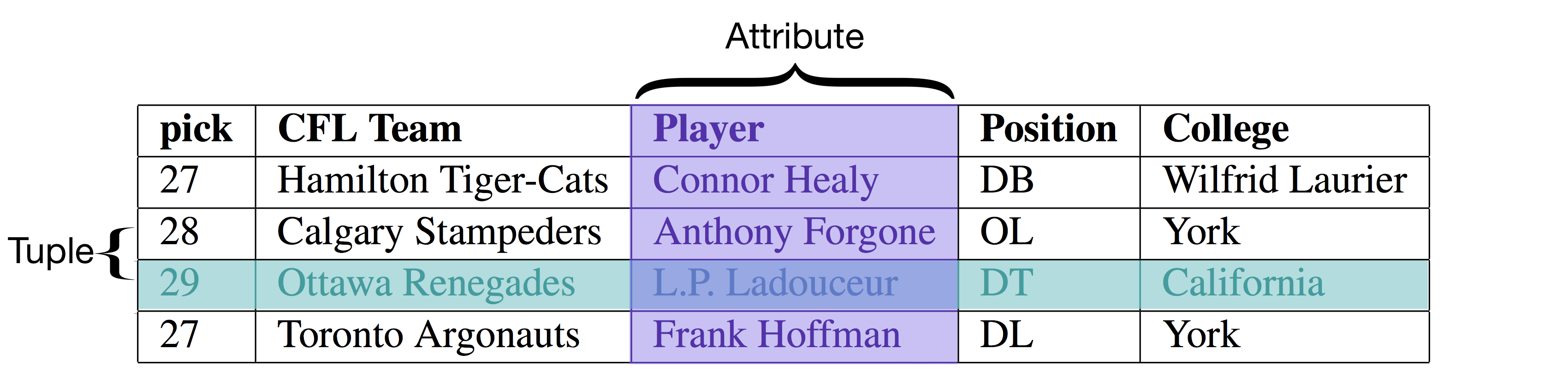}
\caption{Part of a table from the WikiSQL dataset with the contexts within a relation (table) we can model}
\label{fig:table}
\end{figure*}

Relying on the column names is limiting, since the whole model is based on several strong premises: \begin{inparaenum}[(a)] \item the names are high quality and descriptive enough; \item the names do not change; \item the names are known to the user of the bot\end{inparaenum}. These are very strong assumptions: often, column names do not even exist (i.e., the generic col1, col2, etc. are used instead). Moreover, if as we observe in Figure~\ref{fig:table}, a column contains the names of colleges, just changing the column name form "College" to "School" does not make the content any less informative. The expectation from a bot is that their quality is not sensitive to cosmetic changes to the underlying table. Finally, users do not necessarily know the structure of the table, let alone the column names.
%
%
%

In this paper, we build and present ICE (Individual Column Embeddings) -- a novel approach of representing the database table columns, by using their contents instead of their names. To do so, we construct a column embedding vector space, where we embed the columns. This embedding is then used as a substitute for the encoding of the column descriptions (headers) in a state of the art sketch-based model.

In addition, to empirically show the value of using ICE, we generate a new, column-agnostic dataset based on the widely used WikiSQL dataset \cite{DBLP:journals/corr/abs-1709-00103}. In WikiSQL, a substantial bias towards the inclusion in the question of the column name is built-in. For instance more than 79\% of questions contain the name of the column that needs to be selected. Additionally around 59\% contain the names of all columns form the SQL \texttt{where} clause. With ICE, we are eliminating the strong assumption that the users have access to the table structure. Hence, we also need a less biased dataset to show the value of our method.


We thus create an open source data augmentation tool to paraphrase part of the questions in WikiSQL: where the column names are present, we replace them with similar expressions (e.g., synonyms), removing some of the built-in bias.

We train and test our ICE-based model on both the original WikiSQL dataset and our column-agnostic version of the dataset. We show that we maintain the same accuracy on both datasets with all three tasks: aggregation, column-selection and \textit{where} clause generation. We also train the original SQLNet \cite{DBLP:journals/corr/abs-1709-00103} model on the column-agnostic dataset and find a 7\% accuracy drop in the \textit{where} clause generation task.

In a nutshell, the most important contribution of this work is that we \textbf{improve the model resilience} by limiting its reliance on arbitrary descriptions of the data within the tables. In addition, we \textbf{expand the applicability of SQL bots} to users who do not know the internal structure of the databases they are trying to access. By eliminating the need to encode the column headers, we also \textbf{reduce the overall complexity of the model}. This is achieved by removing the LSTM networks used to generate unique column header encodings for the aggregation prediction, selection prediction and \textit{where} clause generation.

The paper is organized as follows: Section~\ref{Related-Work} presents the related work for translating sentences to SQL and for vector space embeddings. In Section~\ref{ICE}, we describe ICE -- our method for column content embeddings. In the next section, we introduce our column-agnostic model for translating sentences to SQL. We present the evaluation results in Section~\ref{Evaluation} and finally conclude in Section~\ref{Conclusion}.

\section{Related Work} 
\label{Related-Work}

\subsection{Related work in Sentences to SQL} 
\label {Related work in Sentences to SQL}

Systems that enable users to use natural language to interact with a database have been researched since the early seventies. As summarized in \cite{DBLP} these early approaches were mostly rule-based. More successful methods have emerged since the advent of the sequence to sequence \cite{DBLP:journals/corr/SutskeverVL14} neural network architectures and increased availability of training data in recent years.  The first model to leverage this was SEQ2SQL introduced by \cite{DBLP:journals/corr/abs-1709-00103} together with their crowdsourced dataset WikiSQL. SEQ2SQL solves the problem of generating SQL queries in a three-step approach that aligns with the structure of an SQL query. 
First, it determines the aggregation function for the query i.e. whether to apply count, average, max etc. This is performed by a classifier trained on the encoding of the question and the encodings of the table headers. 
In the second step, the model determines the column on which to perform the selection, again based on the encoding of the question and the encodings of the table headers. 
Finally, in the last step, the model generates the where clause of the SQL query. To do so it first determines the number of conditions in the clause and then proceeds to generate tuples of a column, comparison operator and value using a pointer network. Since the order in the where clause is not important when there are multiple conditions the model also implements a  reinforcement learning policy to optimize for execution correctness and uses a mixed loss function.

\begin{figure*}[]
\centering
\includegraphics[width=0.75\textwidth]{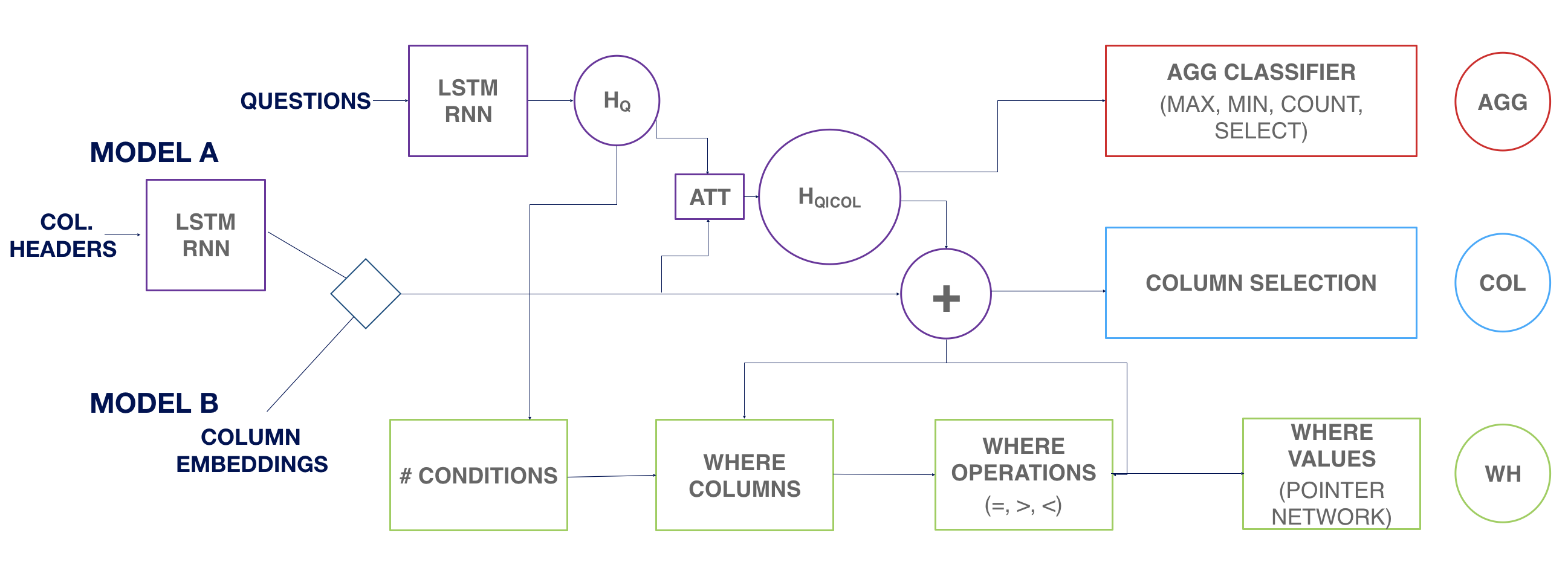}

\caption{The general network architecture of SQLNet. Model A  represents the original model, while Model B represents our model.
}
\label{fig-architecture}
\end{figure*}
 
SQLNet \cite{DBLP:journals/corr/abs-1711-04436} improved upon SEQ2SQL by eliminating the need for reinforcement learning by using a sketch-based approach. \cite{Bornholt:2016:OSM:2837614.2837666, Solar-Lezama:2006:CSF:1168918.1168907} In the where clause section SQLNet introduces a sequence-to-set model. It first picks a set of columns which will be used in the clause. Subsequently, for each column, it determines a comparison operator using a classifier and picks a comparison value using a printer network.  Additionally, this model implements a column attention mechanism which together with sequence-to-set model improves the accuracy over SEQ2SQL by 9\% to 13\%.

\subsection{From Word to Table Embeddings}

The most basic form of word embeddings is the bag of words model. It can be augmented by statistics such as TF-IDF, however, such vector space captures very little of the words semantics, morphology, hierarchy and context. Word2vec, introduced by \cite{NIPS2013_5021} is one of the first popular neural embedding models. It comes in two general implementations: a continuous bag of words (order in window irrelevant) and a continuous skip gram (weight in window based on distance from current word). The objective function of  Word2vec causes words that appear in a similar context to cluster together in the vector space, based on cosine distance. This method was modified by the introduction of global word representation which aims to capture the meaning of the word within the whole corpus \cite{Pennington14glove:global} and the use of subword information to capture the morphology of the words \cite{DBLP:journals/corr/JoulinGBM16}.

With the addition of simple techniques, such as a trained weighted average, word-embedding algorithms were further extended to embed whole sentences \cite{pgj2017unsup} and whole documents \cite{DBLP:journals/corr/LeM14}. Such techniques have also recently been used to get the embedding of whole tables for the purposes of table classification \cite{DBLP:journals/corr/abs-1802-06290}.

\section{ICE: Individual Column Embeddings}
\label{ICE}

To understand the context and the hierarchy of a table we will use the formal definition of a relation: "a set of tuples ($d_1$, $d_2$, ..., $d_n$), where each element $d_j$ is a member of $D_j$, the $j-th$ data domain." Tuples, relations and attributes are graphically depicted in Figure \ref{fig:table}. We observe that there are two contexts in which an element, or cell, $d_j$ appers either within a tuple (row) or within a data domain (column).

To embed the whole table we need to look at both contexts. This complexity is 
not necessary in the context of individual column embeddings, where the latter context is sufficient.
TabVec uses deviation from the median for table vectors to capture the noise \cite{DBLP:journals/corr/abs-1802-06290}, as the final table vector incorporates information from cells that are not conceptually similar. This is not the case for individual column embeddings, as for ICE we assume that the cells within a column are conceptually similar. For instance, if the column is about locations, all the cells are likely to represent location names. This property allows us to simplify the aggregation and use the median vector of all cells as the column representation.

Table cells are not semantic atoms and can contain multiple words, for example in Figure~\ref{fig:table} all \textit{Team} names contain at least two words.
Thus, given a vector space model for words, we compute the individual cell embedding (ICE) as the average of the word embeddings and the individual column embedding as the median of its cells.

To sum up, let a column $D$ contain cells $c_i \in C(D)$, with each cell consisting of a sequence of $n_i$ words ($w_{i1}$, ..., $w_{ij}$, ..., $w_{in_i}$). Given a function $E$ that computes a word embedding, the ICE of the $D$ is defined as:
$$E(D) = median_{c_i}(\frac{1}{n_i} \sum_{j = 0}^{n_i}{E(w_{ij})}), c_i \in C(D)$$

\subsection{Table Word Embeddings.}
For the ICE to be meaningful, the word embeddings need to reflect the table semantics.
The way words are used in tables differs significantly from the way they appear in normal language. We keep the intuition that a word can be represented as an aggretation over all the contexts in which that word appears. What changes from typical text embeddings \cite{NIPS2013_5021, Pennington14glove:global} is that the context is given by other words that occur in the same table column. We view column tables as synthetic sentences that allow us to learn what the relevant context is. We then use SkipGrams with a window of 5 to generate the embedding model.

We first construct a data corpus of synthetic sentences, corresponding to columns.
We define a sentence as all the cells in one table column concatenated. 
Furthermore, we make the assumption that the order of the cells within a column is not important. For the table in Figure~\ref{fig:table}, a sample sentence would be \textit{Calgary Stampeders Ottawa Renegades Toronto Argonauts Hamilton Tiger-Cats}.
We generate 10 random cell shuffles of each column. Using this corpus we train a word2vec model with the Gensim toolkit \cite{rehurek_lrec}.

\section{Individual Column Embedding for Bot Resilience} 
\label{Column-Embedding}
Our work builds upon the SQLNet \cite{DBLP:journals/corr/abs-1711-04436} sketch-based approach. To generate a SQL statement, each component of the query is generated individually: \textit{the aggregation}, the \textit{column selection} and the \textit{where clauses}. The task is thus akin to slot filling \cite{DBLP:journals/corr/abs-1711-04436}. The process is graphically depicted in Figure \ref{fig-architecture}. The input or the SQLNet and previous models \cite{DBLP:journals/corr/abs-1711-04436} consists of a representation of the question and a representation of each table column header. 

We believe this assumption represents one of the most important drawbacks of the approach, as knowledge about the column headers may not exist in real world conditions.
The reason this knowledge was used in previous work is that the dataset itself was biased towards explicitly including the column names in the question formulation. In this section we show how to build a dataset that alleviates this bias. We then use the new dataset to create a model that relies on the column content , not on the column headers.

\subsection{Column-agnostic WikiSQL}

The wikiSQL dataset was crowdsourced using tables from Wikipedia. Workers on Amazon Mechanical Turk\footnote{https://www.mturk.com/} were presented with a table and a generated SQL query and were asked to ask a question that matched the query. This method introduces an inherent bias in the dataset as demonstrated in Table \ref{tab:col_in_qst}. Almost 80\% of questions contain the column name that is retrieved in the selection step and 68\% of questions contain at least one of the column names from the where clause. In total, only 11\% of the questions do not contain \textbf{exact matches} of the column names, as shown in Figure \ref{fig:table}.
As the workers were shown the whole table with column names, in a large number of cases they copied the column name in the question.

\begin{table}
    \centering
    \begin{tabular}{|l|l|l|l|}
\hline
Column type & Train & Test & Dev \\ \hline
Selection col. & 79.0\% & 79.0\% & 79.65\% \\ \hline
Where col. \textgreater{}= 1 & \multicolumn{1}{c|}{68.0\%} & \multicolumn{1}{c|}{67.6\%} & \multicolumn{1}{c|}{68.4\%} \\ \hline
All where col. & \multicolumn{1}{c|}{58.9\%} & \multicolumn{1}{c|}{58.5\%} & \multicolumn{1}{c|}{59.2\%} \\ \hline
\end{tabular}\\
    \caption{The percentages in the table show the proportion  of  questions  that  contain  the  specific  column header in the different data partitions. }
    \label{tab:col_in_qst}
\end{table}

We paraphrase questions that contain a column name to make the dataset more realistic, as described in Algorithm 1. We create candidate questions by replacing the names with synonyms that share the syntactic and semantic properties of the original names.

The original question and the candidate questions are then embedded in vector space with sent2vec \cite{pgj2017unsup}. Using these vector space representations we compute the cosine similarity between the original question and the potential replacements and choose the most similar candidate. This procedure yields a suitable rephrasing for 20\% of the dataset, as we did not find synonyms for all questions containing column names.
For instance, the orginal questions \textit{What is the \textbf{length (miles)} of endpoints westlake/macarthur park to wilshire/western?}, which contains the column header \textbf{length (miles)}, becomes \textit{What is the \textbf {distance (miles)} of endpoints westlake/macarthur park to wilshire/western?}.

\begin{figure}[h]

\centering
\includegraphics[width=0.40\textwidth]{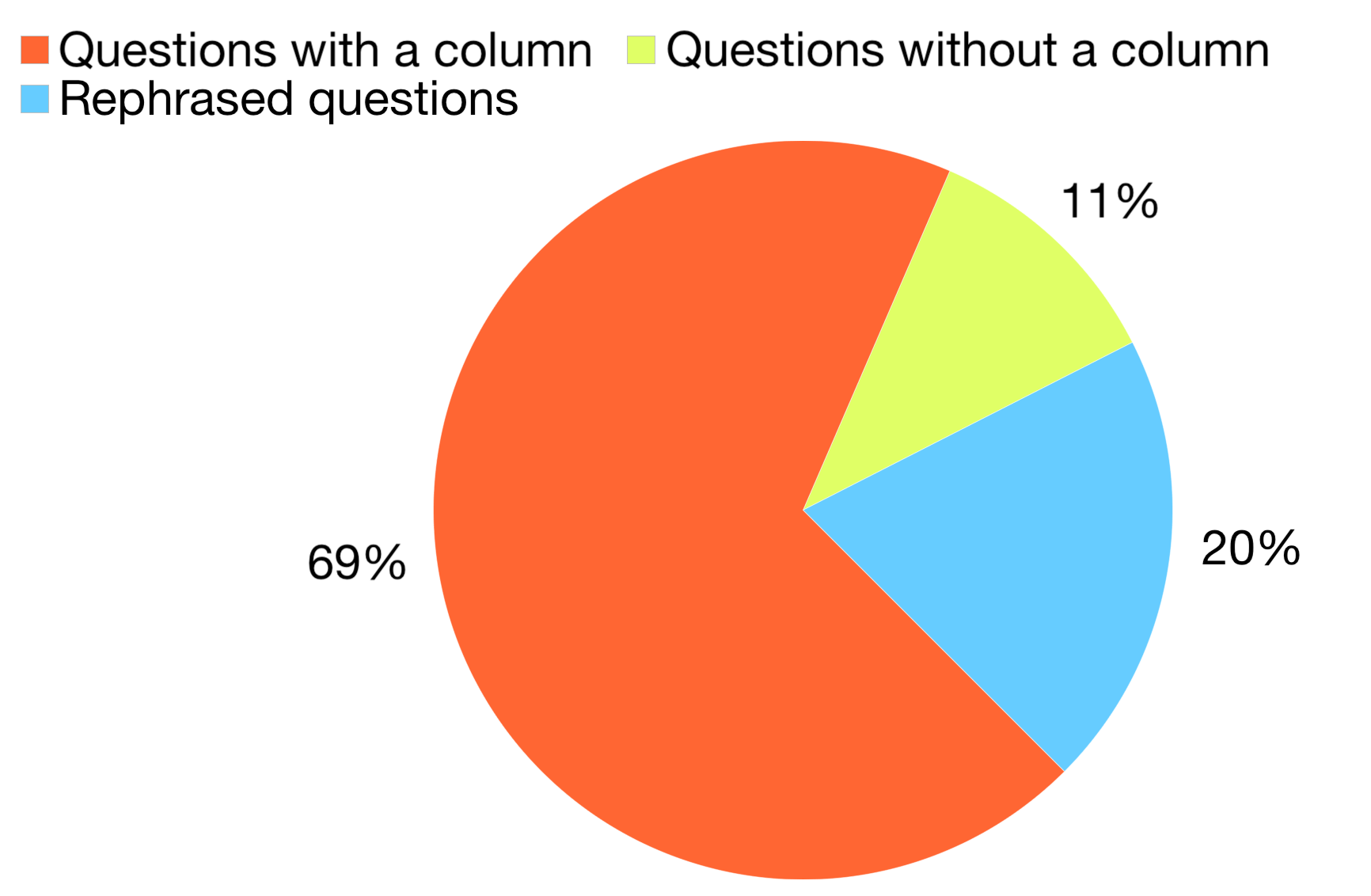}
\caption{Proportions of the modified dataset}
\label{fig:piechart}
\end{figure}

\begin{algorithm}[]
 \KwData{Question and column header}
 \KwResult{Replacement candidate questions}
 Tokenize and pos tag question\;
 \For{word in column header}{
  Get word tag in question\;
  Get word synonyms using tag\;
  \If{synonyms list \textgreater{} 0}{
   append synonyms to rephrase list\;
   }
 }
 \For{phrase in rephrase list}{
  \If{length of phrase ==  length of header}{
   replace column header in question with phrase\;
   append new question to candidate list\;
   }
 }
 
 \caption{Generating replacement candidate questions}
\end{algorithm}

\subsection{Integrating Individual Column Embeddings} 

We compute the embeddings for the entire table column corpus as described in chapter \ref{ICE}. This is necessary since the embeddings are required during inference both during training and testing. Due to model size constraints, we keep the individual column embeddings constant during both training and testing. 
We create a dictionary to link each column to its embedding vector and feed it to the model (Model B) in Figure \ref{fig-architecture}. An attention mechanism has the embeddings as inputs and the result contributes to the aggregation, selection and where clause modules. The column vectors are generated with the same dimensions that we use for the question encoding.

As we replace the column headers with column content embeddings, our model is completely agnostic to the headers. We thus remove the LSTM used to encode the column headers in the three model components: aggregation, selection and where clause generation. This leads to a significant \textbf{reduction in the complexity of the model}.

\section{Evaluation}
\label{Evaluation}

\begin{table*}
  
    \begin{tabular}{|l|c|c|c|c|c|c|}
\hline
 & \multicolumn{3}{l|}{Dev Set Accuracy} & \multicolumn{3}{l|}{Test Set Accuracy} \\ \hline
 & \multicolumn{1}{l|}{Aggregation} &  \multicolumn{1}{l|}{Selection} & \multicolumn{1}{l|}{Where-clause} &  \multicolumn{1}{l|}{Aggregation} &  \multicolumn{1}{l|}{Selection} &  \multicolumn{1}{l|}{Where-clause} \\ \hline
Seq2SQL & 90.0\% & 89.6\% & 62.1\% & 90.1\% & 88.9\% & 60.2\% \\
SQLNet & 90.1\% & 91.5\% & 74.1\% & 90.3\% & 90.9\% & 71.9\% \\
SQLNet + ICE & 89.7 \% & 92.4 & 72.2\% & 89.3 \% & 91.8 & 71.1\% \\ \hline
\end{tabular}
    \caption{
Model accuracies on the Original WikiSQL Dataset}
    \label{tab:res1}
\end{table*}

\begin{table*}[]
  
    \begin{tabular}{|l|c|c|c|c|c|c|}
\hline
 & \multicolumn{3}{l|}{Dev Set Accuracy} & \multicolumn{3}{l|}{Test Set Accuracy} \\ \hline
 & \multicolumn{1}{l|}{Aggregation} &  \multicolumn{1}{l|}{Selection} & \multicolumn{1}{l|}{Where-clause} &  \multicolumn{1}{l|}{Aggregation} &  \multicolumn{1}{l|}{Selection} &  \multicolumn{1}{l|}{Where-clause} \\ \hline

SQLNet & 90.1\% & 87.5\% & 63.4\% & 90.3\% & 87.1\% & 63.1\% \\
SQLNet + ICE & 89.7 \% & 88.4 & 70.1\% & 89.3 \% & 87.9 & 69.4\% \\ \hline
\end{tabular}
    \caption{Model accuracies on the Column-agnostic WikiSQL Dataset}
    \label{tab:res2}
\end{table*}

\subsection{Original WikiSQL Evaluation}
The evaluation on the full original WikiSQL dataset determines whether the individual column embeddings are suitable replacements for headers when the column name appears in the question.
Table \ref{tab:res1} summarizes the results of our model \textit{SQLNet+ICE} and compares them with the results of two baselines: \textit{SQLNet}  and \textit{Seq2SQL}. We portray the accuracy values on the development and test sets for the three slots we fill in the sketch: \textit{Aggregation function}, \textit{Column Selection} and \textit{Where clause generation}. 

We observe that \textit{SQLNet+ICE} performs similarly to the original \textit{SQLNet} model in both cases and superior to \textit{Seq2SQL}. This result shows that we can build an equally performing model that is resilient to changes to the DB schema or complete absence of knowledge about it. 

We note that the accuracy of the aggregation function also changes. This happens because the aggregation classifier has either the column or header embeddings as inputs, as shown in \ref{fig-architecture}.
There is a small decrease of accuracy for the Aggregation and Where clauses, while the accuracy on the Column Selection performs slightly better. These results are expected, as the queries strongly rely on the direct column names mentions.

\subsection{Column-agnostic WikiSQL Evaluation}
The second experiment shows the more realistic results, obtained on the column-agnostic WikiSQL Dataset. 
The results in Table \ref{tab:res2} show that SQLNet struggles to predict correctly the column related parts of the query, especially in the case of the where clause generation.
This drop in the accuracy is expected, since the where clause predictor is the most complex part of the model. 
Without the original dataset bias where the column names were present in the questions, the column names are not descriptive enough.
This leads to a drop of 10.7\% on the validation and 8.8\% on test dataset.  

On the other hand, our model is capable of overcoming this situation and find the queries with a much smaller drop of accuracy. Although the performance is also worse than with the original dataset, the accuracy obtained using SQLNet with individual column embeddings in the where clauses is only 2.1\% lower in validation and 1.7\% in test. Using individual column embeddings makes the SQLNet model more versatile, as it can address the scenario where the user is not aware of the table structure.

\begin{table}[]
  
    \begin{tabular}{|l|c|c|c|}
\hline
 &   \multicolumn{3}{l|}{Rephrased Test Set Accuracy} \\ \hline
 & \multicolumn{1}{l|}{Agg.} &  \multicolumn{1}{l|}{Sel.} &  \multicolumn{1}{l|}{W.-clause} \\ \hline

SQLNet &  89.5 \% & 81.3\% & 43.2\% \\
SQLNet + ICE  & 88.9 \% & 83.2 & 61.3\% \\ \hline
\end{tabular}
    \caption{Model accuracies on the paraphrased questions only on Aggreation, Selection and Where-clause tasks.}
    \label{tab:res3}
\end{table}

\textbf{Focusing on rephrased questions.}
To better understand our results on the Column-agnostic WikiSQL dataset we run the evaluation just with questions that have been rephrased, which represent around 20\% of the whole data set, as shown in Figure \ref{fig:piechart}. Table \ref{tab:res3} summarizes these results, with SQLNet is the original model described in \cite{DBLP:journals/corr/abs-1711-04436}. The previously seen drop in \textit{SQLNet} accuracy on the column selection and where-clause predictions is exacerbated - showing that indeed the paraphrasing is indeed the root cause. This effect is comparatively mild in \textit{SQLNet + ICE}.

\section{Conclusion and Future Work}
\label{Conclusion}

In this paper, we proposed a new approach to build SQL chatbots without relying on the database table schema. 
Previous work built around the WikiSQL dataset take advantage of the dataset biases and use the column names to improve performance. This reliance on the schema inhibits their generalization capacity to cases where schema knowledge is absent.
Our model, built on SQLNet by adding Individual Column Embeddings \textit{SQLNet + ICE}, does not suffer from these limitations. 

We provide a way to create Individual Column Embeddings, different from the Column Embeddings in prior art \cite{DBLP:journals/corr/abs-1802-06290}. Furthermore, we publish a method to paraphrase WikiSQL questions to alleviate the dataset bias.

The results of our model on the paraphrased WikiSQL are very similar to the ones obtained on the original dataset, while the SQLNet models struggles to deal with the paraphrasing.

\textbf{Future Work.} Even with these changes, there is still room for improvement in the SQL chatbot area. Large scale operations need the support for multiple tables at the time as well as more operations such as \textit{join}. While WikiSQL is a good starting point and our modified version removes some of the biases present in it, there is a strong need for more data, both in terms of quantity and diversity. This new data needs to include more operations, as well as new ways to collect questions to have more variety in the structure of the user's utterances.

\end{document}